\documentclass[10pt,conference,compsocconf]{IEEEtran}
\usepackage{pgfplots}
\usepackage{pgfplotstable}
\pgfplotsset{
    myplotstyle/.style={
    legend style={draw=none, font=\small},
    legend cell align=left,
    legend pos=north east,
    ylabel style={align=center, font=\bfseries\boldmath},
    xlabel style={align=center, font=\bfseries\boldmath},
    x tick label style={font=\bfseries\boldmath},
    y tick label style={font=\bfseries\boldmath},
    scaled ticks=false,
    every axis plot/.append style={thick},
    },
    compat=1.18
}

\graphicspath{ {./figures/} }

\usepackage[utf8]{inputenc}
\usepackage[letterpaper, margin=2.5cm]{geometry}

\usepackage{amsmath}
\usepackage{amssymb}
\usepackage{bm}                     

\usepackage{pdfpages}

\usepackage{wrapfig}

\usepackage{xfrac}

\usepackage{caption}
\usepackage{subcaption}

\usepackage{comment}

\usepackage{fancyhdr}

\usepackage{url}
\usepackage{hyperref}
\hypersetup{
  colorlinks,
  citecolor=blue,
  linkcolor=black,
  urlcolor=[RGB]{255 157 39}}
\usepackage[acronym,nomain,nonumberlist]{glossaries}
\usepackage{glossary-longragged}
\makeglossaries
\newglossarystyle{super3colleft}{%
    {\tablehead{}\tabletail{}%
     \begin{supertabular}{@{}>{\bfseries}lp{\glsdescwidth}p{\glspagelistwidth}}}%
    {\end{supertabular}}%
  \renewcommand*{\glsgroupheading}[1]{}%
}

\usepackage{color,soul}
\usepackage{authblk}
\usepackage{float}
\usepackage{graphicx}               
\usepackage{gensymb}                
\usepackage{graphbox}
\usepackage{capt-of}
\usepackage{xparse}                 
\usepackage{hyperref}
\usepackage{graphicx}   

\makeatletter
\newcommand*\titleheader[1]{\gdef\@titleheader{#1}}
\AtBeginDocument{%
    \let\st@red@title\@title
    \def\@title{%
        \bgroup\normalfont\large\centering\@titleheader\par\egroup
        \vskip1.5em\st@red@title}
}
\makeatother

\titleheader{IAC-24-A3/IP/83}
\title{
    \vspace*{-0.5cm} 
    \Large \textbf{COFFEE: A Shadow-Resilient Real-Time Pose Estimator for Unknown Tumbling Asteroids using Sparse Neural Networks} 
    \vspace*{-0.25cm}
}

\author{\textbf{
Arion Zimmermann$ ^{1} $, Soon-Jo Chung$ ^{2}$, Fred Hadaegh$ ^{2}$} \\\\
  \textit{$ ^{1} $Department of Electronic and Electrical Engineering, EPFL, Lausanne, Switzerland}\\
  \textit{$ ^{2} $Division of Engineering and Applied Science, California Institute of Technology, Pasadena, USA}
}

\begin{document}

\twocolumn[
\begin{@twocolumnfalse}
\maketitle
\end{@twocolumnfalse}
\begin{abstract}
The accurate state estimation of unknown bodies in space is a critical challenge with applications ranging from the tracking of space debris to the shape estimation of small bodies. A necessary enabler to this capability is to find and track features on a continuous stream of images. 
Existing methods, such as SIFT, ORB and AKAZE, achieve real-time but inaccurate pose estimates, whereas modern deep learning methods yield higher quality features at the cost of more demanding computational resources which might not be available on space-qualified hardware. 
Additionally, both classical and data-driven methods are not robust to the highly opaque self-cast shadows on the object of interest. We show that, as the target body rotates, these shadows may lead to large biases in the resulting pose estimates. For these objects, a bias in the real-time pose estimation algorithm may mislead the spacecraft's state estimator and cause a mission failure, especially if the body undergoes a chaotic tumbling motion.
We present COFFEE, the Celestial Occlusion Fast FEature Extractor, a real-time pose estimation framework for asteroids designed to leverage prior information on the sun phase angle given by sun-tracking sensors commonly available onboard spacecraft. By associating salient contours to their projected shadows, a sparse set of features are detected, invariant to the motion of the shadows. A Sparse Neural Network followed by an attention-based Graph Neural Network feature matching model are then jointly trained to provide a set of correspondences between successive frames. The resulting pose estimation pipeline is found to be bias-free, more accurate than classical pose estimation pipelines and an order of magnitude faster than other state-of-the-art deep learning pipelines on synthetic data as well as on renderings of the tumbling asteroid Apophis.\end{abstract}
\bigskip]

\section{Introduction}

In 2004, astronomers Roy Tucker, David Tholen and Fabrizio Bernardi discovered the Apophis asteroid during its first observable fly-by of Earth \cite{chesley2005}. Apophis was long considered as one of the most threatening asteroids due to its proximity to Earth and the uncertainty of its trajectory. Recent observations \cite{pravec2014} \cite{brozowic2022} fortunately contradicted this hypothesis but highlighted the interesting tumbling motion of this asteroid.

Tumbling is a state of motion in which a rigid body with a non-uniform inertia matrix starts rotating around its second principal axis. A small perturbation in its motion then drives the body into a chaotic rotation, known as the Dzhanibekov effect. Dissipative forces limits the time span of this effect on Earth but not in space, where the body may tumble indefinitely.

In 2029, the Apophis asteroid will undergo yet another fly-by of Earth and many space agencies have already planned missions to study this asteroid in detail, such as the OSIRIS-APEX \cite{giustina2023} from NASA and Ramses \cite{kueppers2023} from ESA.

Throughout this paper, we will refer to Apophis as the typical example of a tumbling asteroid, as it is of high scientific interest. Nevertheless, all methods shown in this paper are more generally applicable to tumbling objects with an unpredictable motion, be they small bodies, space debris, satellites or other unknown objects.

Some missions require spacecraft to synchronize their orbit to the motion of the asteroid before engaging a final approach to its surface. Typically, ground operators would collect imagery data from the spacecraft and deduce the long-term motion of the asteroid by manually matching highly descriptive features on its surface, such as the edge of craters and the shape of boulders. Photogrammetry or Stereophotoclinometry (SPC) methods would be used to obtain the shape model of the asteroid and an accurate estimation of its motion. Up to now, missions to asteroids have always included a significant portion of their mission time to analyze the target asteroid's trajectory before attempting an approach.

In the case of Apophis, this method cannot be used to predict the pose of the asteroid since the instability in its rotation leads to an exponential prediction error, even for a short time horizon. This motivates the need to develop an autonomous algorithm to estimate the pose of the asteroid in real-time.

In addition to this, asteroids have a highly uniform surface material. As a result, the number of visible features is drastically reduced compared to an indoor or outdoor environment on Earth. All visible features on asteroids are in fact the result of the self-cast shadows of boulders and craters that lie on their surface. As the asteroid rotates, the highly opaque self-cast shadows introduce a bias in the instantaneous pose estimates, because the shadows are distorted and move differently than the asteroid itself. 

The purpose of this paper is to design a robust and accurate real-time algorithm capable of estimating the instantaneous pose of a tumbling small body, thereby enabling orbit synchronization, which is a key capability for potential scientific missions around tumbling bodies in space. 

\section{Related work}
The real-time pose estimation of asteroids usually involves classical feature detectors, descriptors and matchers to find keypoint correspondences between successive frames. The detected features are salient points processed by handcrafted detectors and descriptors such as SIFT \cite{sift}, ORB \cite{orb} or AKAZE \cite{akaze}. Well-researched methods \cite{nesnas2021} exist to process such handcrafted feature in an autonomous navigation framework around asteroids. 

Several studies also focused on the detection and tracking of natural landmarks, such as craters or boulders \cite{chen2021} \cite{leroy2001}. 
The relative pose between the spacecraft and the asteroid is then either formulated as a Maximum-Likelihood estimation problem to minimize the reprojection error between multiple frames \cite{takeishi2015} or as a pose-graph optimization problem \cite{dor2021} \cite{nakath2020}. 

In all these cases, large and opaque shadows can mislead the pose estimation algorithm to yield erroneous measurements because of the inherent lack of robustness from handcrafted feature descriptors.

To improve the robustness of these algorithm to different lighting conditions, deep learning methods have been developed, which leverage Convolutional Neural Networks and Multi-Layer Perceptrons. Some methods \cite{sharma2018} \cite{park2024} made use of a fully learned pipeline to estimate the pose of an uncooperative spacecraft directly and others \cite{kaluthantrige2023} used a CNN-based approach to regress the relative pose between a camera and an asteroid. Most of these deep learning methods are nevertheless limited by the current computing capabilities of space hardware and cannot run in real-time.

Some of these studies however acknowledged the complex effect of moving shadows due to a varying sun phase angle and asteroid rotation.  In \cite{nesnas2021}, this effect, called the "feature drift", was modelled by a constant angular velocity model for each feature in the asteroid's reference frame. For tumbling bodies, this motion model cannot be used and the feature drift must be compensated on a frame-to-frame basis.

In contrast to current deep learning methods, we will propose a lightweight real-time deep learning algorithm to estimate the pose of the target asteroid, while avoiding the feature drift bias caused by moving shadows.

Even though deep learning improved the invariance of the pose estimation algorithm to changing lighting conditions, only a few methods have explicitly incorporated the sun phase angle as a prior to the state estimation pipeline. \cite{chen2024} used Neural Radiance Field to find a dense solution to represent the shape of an asteroid. Even though not real-time, this method is the first to incorporate information about the camera and the sun phase angle as an input embedding to their model. 

We will show that treating the sun phase angle as a prior can not only improve the accuracy of deep learning pipelines but also significantly accelerate them. 

\section{Problem statement}
We focus on the proximity operations of a spacecraft in the vicinity of a tumbling asteroid. The spacecraft is assumed to have limited computational resources, a single monocular camera, a ranging sensor, a sun-tracker and a star-tracker. 

We want to design a real-time pose estimation algorithm which allows the spacecraft to synchronize its orbit with respect to the tumbling asteroid for an imminent landing or touch-and-go operation. Since the asteroid is tumbling, this algorithm must be robust to large viewpoint changes (\textgreater 10°) between successive frames. In addition to this, it must be robust to the feature drift of moving shadows and exhibit a low pose estimation bias.

\begin{figure}[!htb] 
\centering
\includegraphics[width=0.5\textwidth, trim=0 0cm 0 0cm, clip]{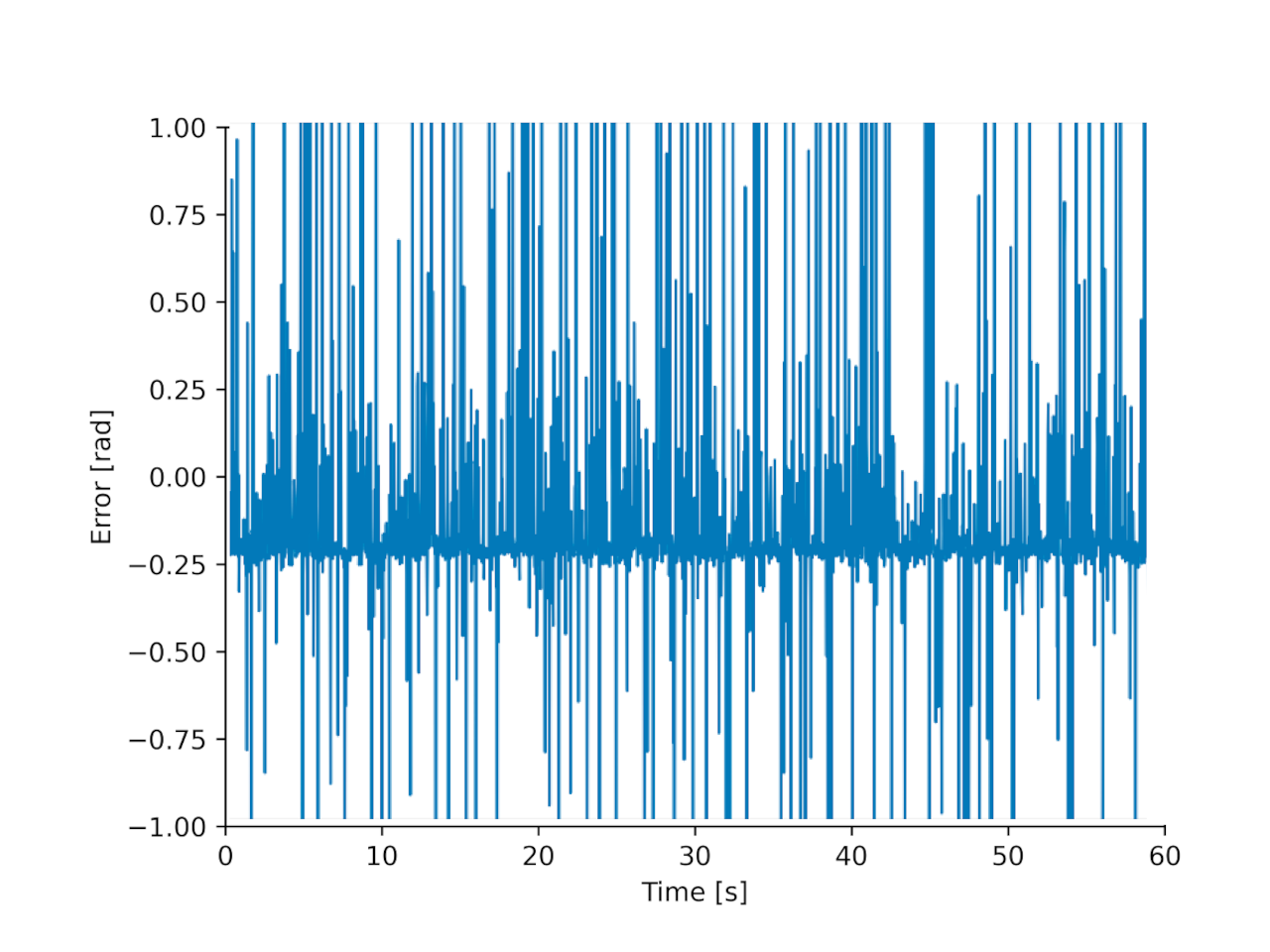} 
\caption{Pose estimation bias due to moving shadows when using SIFT on an asteroid with few visible features} 
\label{fig:bias} 
\end{figure}

\section{Method}

The COFFEE (Celestial Occlusion Fast FEature Extractor) pose estimation pipeline consists of five steps. An image is first captured from the onboard camera sensor and the sun phase angle is measured from the sun-tracker. This data is used to detect a sparse set of relevant keypoints on the image. A learned feature descriptor then assigns a feature vector to each of the detected keypoints. As multiple camera frames and sun phase angles are captured, a data-driven matching network finds feature correspondences between subsequent frames. Finally, a robust scheme on top of an analytic pose solver estimates the relative pose from the found correspondences. 

\subsection{Keypoint Detection}

Extensive research has already been done on the design of keypoint detectors, yet most of the current algorithms rely on the assumption that edges or corners with high contrast are likely to reappear in subsequent frames at the same location relative to one another. This assumption is well verified in indoor and outdoor environments on Earth and has been used in a variety of computer vision algorithms.

However, this assumption does not hold for rotating small bodies in space, which have salient boulders and craters on top of a highly uniform surface. All high contrast edges and corners are in fact the self-cast shadows of these salient features. In the case of a rotating body, the shadows move inconsistently with respect to the actual geometry of the body.

Current feature detectors fail to recognize which edges and corners are part of the geometry of the asteroid and which are self-cast shadows of this geometry. This shortcoming may lead to a strong bias in the pose estimation of the asteroid for some distribution of boulders and craters. Figure \ref{fig:bias} shows this bias when applying the SIFT feature descriptor on renderings of a rotating synthetic asteroid with little surface roughness. The plotted pose is estimated with a RANSAC scheme on top of a 5-point pose solver. The SIFT algorithm exhibits a strong bias of -0.18 rad because it is not robust to the motion of the opaque self-cast shadows on the target asteroid.

Here, we propose a keypoint detection algorithm that leverages the data from the onboard sun-tracker to have a priori information about which parts of the image are self-cast shadows and which parts are not. 

Let’s define the following reference frames:
\begin{enumerate} 
\item Inertial reference frame $F_i$ 
\item Asteroid reference frame $F_a$
\item Camera reference frame $F_c$
\end{enumerate}

Let the unit vectors $\mathbf{D}_i$ be the direction of the sun rays in the $F_i$ frame and $\mathbf{D}_c$ be the direction of the sun rays in the $F_c$ frame. $-\mathbf{D}_c$ is given by the sun-tracker. 

All $k$ points on the asteroid can be represented by $\mathbf{P}_i^{(k)}$ in the inertial reference frame.

A ray is traced from each of point $\mathbf{P}_i^{(k)}$ in the direction of the sun vector $\mathbf{D}_i$, until it hits again the geometry of the asteroid. All such lines obey to equation \eqref{eq:rays} for any scalar $\lambda \in \mathbb{R}$.

\begin{equation}
\label{eq:rays}
\mathbf{Z}_i^{(k)}(\lambda) = \mathbf{P}_i^{(k)} + \lambda \mathbf{D}_i
\end{equation}

Under the directional light source assumption, all casted rays are parallel to each other, implying that their projection onto the camera image plane is rendered as lines that intersect themselves into a vanishing point. Indeed, the points $\mathbf{Z}_i^{(k)}(\lambda)$ can be transformed in the camera's reference frame through the rotation matrix $\mathbf{R}^i_c$, such that
$\mathbf{Z}_c^{(k)}(\lambda) = \mathbf{R}^i_c \mathbf{Z}_i^{(k)}(\lambda)$

We assume that $\mathbf{R}^i_c$ si given by the spacecraft’s state estimator and is determined thanks to the onboard sun-tracker and star-tracker.

The vanishing point $v$ can now be expressed by taking the limit as $\lambda \rightarrow \infty$ and applying the perspective projection $Proj$ which divides all coordinates by their third coordinate (depth):

$$
v = \lim_{\lambda \rightarrow \infty} Proj(\mathbf{Z}_c^{(k)}(\lambda)) = 
\begin{bmatrix}
\mathbf{D}_c^T \mathbf{e}_x  / \mathbf{D}_c^T \mathbf{e}_z \\
\mathbf{D}_c^T \mathbf{e}_y / \mathbf{D}_c^T \mathbf{e}_z \\
\end{bmatrix}
,$$ where $\mathbf{e}_x$, $\mathbf{e}_y$, $\mathbf{e}_z$ is the basis of the $F_c$ frame expressed in the $F_c$ frame.

The true pixel location of the vanishing point is finally given by $v’ = Kv$, where $K$ is the intrinsic matrix of the camera.

\begin{figure}[!htb] 
\centering
\includegraphics[width=0.5\textwidth, trim=0 0cm 0 0cm, clip]{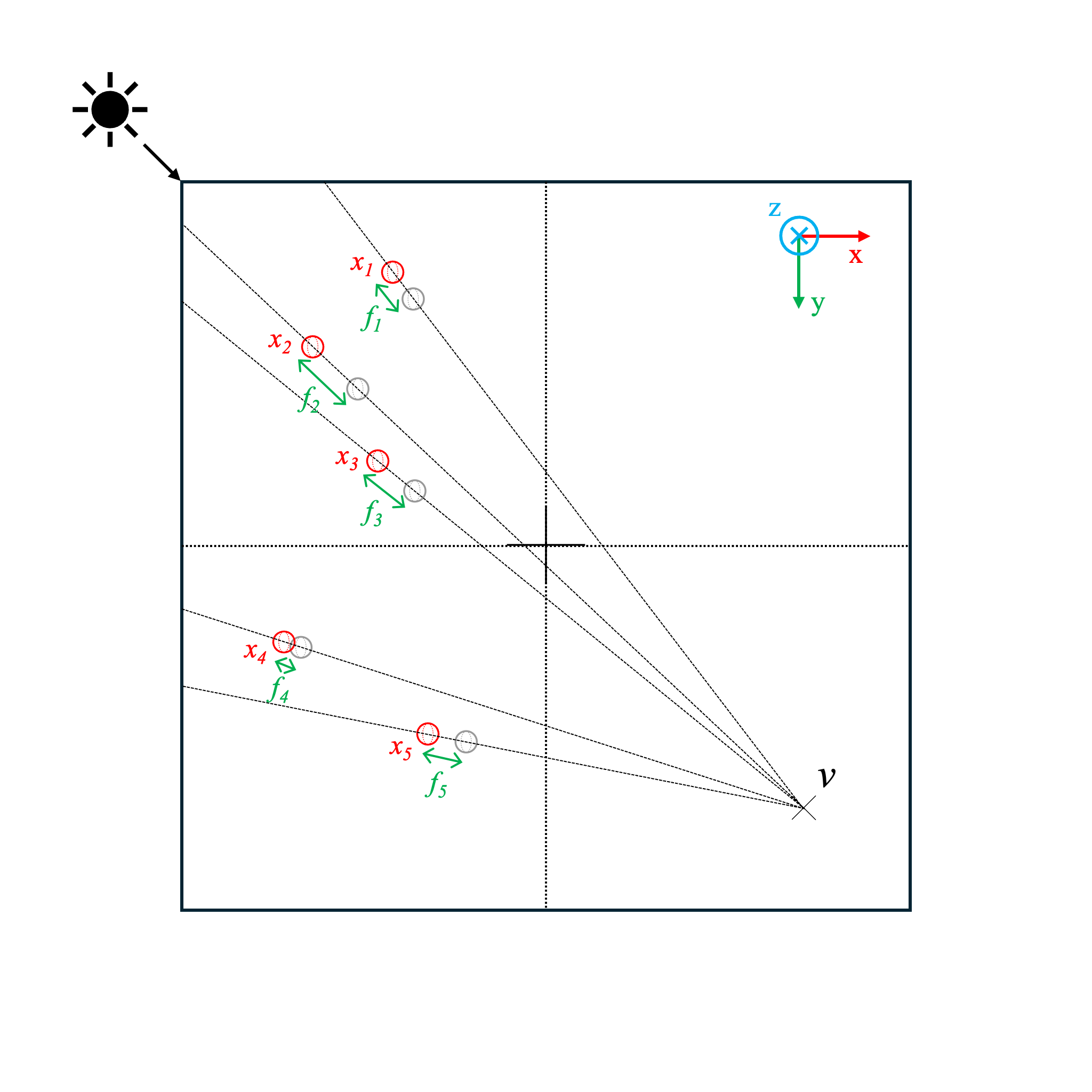} 
\caption{Projection of light rays to the vanishing point $v$, keypoint selection (red) and shadow size encoding (green)} 
\label{fig:lines3d} 
\end{figure}

Figure \ref{fig:lines3d} illustrates this phenomenon by depicting the projection of light rays and their vanishing point.

All points belonging to the geometry of the asteroid may only cast a shadow in the direction of the vanishing point. Thereby, we can design an algorithm that takes advantage of this fact to distinguish between features and their cast shadow. 

The benefits of this method are twofold:
\begin{enumerate}
\item It allows us to find salient features on the asteroid (e.g. craters and boulders) that will likely remain illuminated through time.
\item It prevents us from selecting keypoints at the edge of the shadow rays which would bias the resulting pose estimates.
\end{enumerate}
An edge filter is applied along the rays vanishing at $v’$ and a sparse representation is extracted by only keeping the coordinate locations of the negative edges and encoding the size of the shadows (pixel distance between the negative edge and the next positive edge) as the features corresponding to the said coordinate locations, as shown in figure \ref{fig:lines3d}. With this method, the detected keypoints are not only robust to shadows but also incorporate the size of their cast shadow, as additional information for the description and matching stages.

\subsection{Feature Description}
The feature detection algorithm generates a set of keypoints that are anchored to the geometry of the asteroid. One floating-point value representing the size of the cast shadow is assigned to each of these points.  

The purpose of the feature description algorithm is to transform the shadow size feature into a high dimensional space which also describes the relation of a given keypoint to its neighbors. Doing so will allow the matching algorithm to use information about the geometry of the mesh and the shape of the craters and boulders to infer which features are matching with one another. 

State-of-the-art data-driven feature descriptors use Convolutional Neural Networks (CNNs) to augment the dimensionality of the input image and extract relevant features from it. These deep learning layers are used in an encoder/decoder architecture, where features are extracted and merged at different scales from the input image, before being restored to their original scale on a higher-dimensional space. 

Our feature description algorithm exploits the unique properties of the detected keypoints. We observe that these are derived from the projection of the geometry of the asteroid onto itself through its self-cast shadow. We encoded the boundaries of this shadow in the detected features. Since the geometry of the asteroid is assumed to be made from a continuous mesh, its projection is a continuous shape and the detected keypoints form a continuous curve.

Instead of describing the input image around these selected keypoints, like most feature description algorithms do, we leverage the remaining continuity structure of keypoint distribution to create the feature vector. This has the advantage of not requiring to process the data from the whole image again. Nonetheless, it assumes that enough information is stored in the distribution of the keypoints to match the features with one another. 

The structure of spatially continuous features is typically extracted through nested CNN layers. Notwithstanding, using CNNs directly on the keypoints would waste key computational resources since only a very small subset of the image space is filled by keypoints (sparsity ratio ca. 1:500).

The keypoints are consequently processed by a Sparse Submanifold CNN \cite{graham2018} which convolves the features on the sparse space of keypoints. Hash tables and precomputed mappings between the layer's inputs and outputs makes the sparse submanifold convolutions extremely efficient on modern GPUs. 

Our CNN architecture is composed of four ResNet \cite{resnet} bottleneck blocks in the encoder stage due to their fast inference time and their ability to extract contextual information from the sparse set of keypoints. Layers of transpose convolutions with batch normalization are used in the decoder stage. The resulting network is composed of 17 Sparse CNN layers, as described in the Appendix.

\subsection{Feature Matching}

One of the weaknesses of submanifold CNNs is that the filter kernels still lie on the image space, as opposed to the data lying in the submanifold space. This prevents the CNN from efficiently learning relations between the extracted curves, without prohibitively increasing the size of the network’s receptive field.

We therefore augment the resulting feature vectors by encoding the keypoint coordinates to them. An attention-based Graph Neural Network (GNN), based on the Lightglue \cite{lightglue} architecture is then used to encode relations between features within the input image (self-attention) and across multiple frames (cross-attention). 

A GNN expresses the geometrical links between the two sets of image points. An undirected complete graph $\mathcal{G}$ is created, where each node represents a keypoint. The edges connecting points from image A to image A and from image B to image B are called self-edges $\varepsilon_\text{self}$ and the edges connecting points from image A to image B are called cross-edges $\varepsilon_\text{cross}$.  

After encoding the keypoint location with the descriptor value, a multi-head attention layer is applied separately to $\varepsilon_\text{self}$ and $\varepsilon_\text{cross}$, so that the state of each node in the graph can be updated accordingly through message passing \cite{gnn}. Multiple layers of successive self-attention and cross-attention are used to build a faithful latent representation of the geometry of the underlying scene. 

\subsection{Pose Estimation}
\label{poseest}
Given a set of correspondences between keypoints, it is possible to recover the relative pose between the camera reference frame $F_c$ and the asteroid reference frame $F_a$.  

This problem is usually formulated as solving the camera pose with respect to a fixed reference frame, which in our case is the asteroid reference frame. Note that the resulting pose estimate has its translation component always normalized, since it is not possible to know the distance to the asteroid with a single monocular camera. The onboard ranging sensor is used to obtain this depth information. 

Any pair of corresponding features in two views of the asteroid satisfy the epipolar constraint \eqref{eq:epipolar}, encoding the fact that each point in 2D from one view must lie on the so-called epipolar line in the other view.
\begin{equation} 
\label{eq:epipolar}
\mathbf{x'}^\top \mathbf{F} \mathbf{x} = 0
\end{equation}

We assume that we know the camera's intrinsic parameters $K$ and can use this information to express the fundamental matrix $F$ in terms of the essential matrix $E$, such that

\begin{equation} 
\mathbf{F} = \mathbf{K}^{-\top} \mathbf{E} \mathbf{K}^{-1} 
\end{equation}

Moreover, the essential matrix must have a rank of two \eqref{eq:essconst1} and have the same non-zero singular values \eqref{eq:essconst2}. 

\begin{equation} 
\label{eq:essconst1}
\det(\mathbf{E}) = 0
\end{equation} 

\begin{equation} 
\label{eq:essconst2}
2 \mathbf{E} \mathbf{E}^\top \mathbf{E} - \text{trace}(\mathbf{E} \mathbf{E}^\top) \mathbf{E} = 0
\end{equation}

The 5-point algorithm \cite{nister2005} is used to find the relative pose of the asteroid by solving \eqref{eq:epipolar} under the constraints \eqref{eq:essconst1} and \eqref{eq:essconst2} up to a scale ambiguity. \\

The relative pose ($\mathbf{R}$, $\mathbf{t}$) between is then easily recovered through singular value decomposition since
\begin{equation} \mathbf{E} = [\mathbf{t}]_\times \mathbf{R} \end{equation} 
where \begin{equation} [\mathbf{t}]_\times = \begin{pmatrix} 0 & -t_z & t_y \\ t_z & 0 & -t_x \\ -t_y & t_x & 0 \end{pmatrix} \end{equation} is the skew-symmetric matrix representation of the translation vector.

A RANSAC scheme \cite{ransac} is employed on top of the 5-points algorithm to remove outliers from the set of matches. 

The full COFFEE architecture is summarized in the Appendix.

\section{Dataset generation}
Asteroids observed on recent missions always exhibited low and stable rotation rates, preventing us from using their imagery data to train, validate or test the COFFEE algorithm.  We therefore developed a synthetic dataset of multiple asteroids viewed under different viewpoints. 

We first created shape models of asteroids by either enhancing existing shape models or applying a generative procedure. A physics simulation software was then used to render realistic frames of the small bodies on a tumbling motion.

For the train, validation and test datasets, Blender and MONET \cite{monet} were used to generatively create shape models. Blender is a well-known rendering software, mostly used to create realistic animations. MONET is a software enhancing shape models of asteroids by adding crater impacts, boulders and roughness to their surface. Note that MONET uses normal maps to render the surface roughness. Rendering with normal maps, as opposed to fully ray-traced rendering, does not allow for ray-traced shadows, which are crucial for COFFEE. We hence reimplemented MONET to transform the mesh of the asteroid instead of using normal maps. All parameters pertaining to the generation of the shape models were drawn from the distributions defined in table \ref{table:dsgen_summary}. 
\begin{table}[!htb] 
\centering 
\begin{tabular}{|c|c|} 
\hline \textbf{Parameter} & \textbf{Value} \\ 
\hline Roughness & $U(2, 10)$ \\ 
\hline Scale factor & $U^3(1, 3)$ \\ 
\hline Deform & $U(1, 10)$ \\ 
\hline Depth & $U(0.1, 0.5)$ \\ 
\hline Large rock count & $U(1, 10)$ \\ 
\hline Medium rock count & $U(10, 100)$ \\ 
\hline Small rock count & $U(100, 1000)$ \\ 
\hline Large rock size & $U(0.01, 0.03)$ \\ 
\hline Medium rock size & $U(0.003, 0.01)$ \\ 
\hline Small rock size & $U(0.001, 0.003)$ \\ 
\hline \end{tabular} 
\caption{Parameter distributions for the generation of shape models} 
\label{table:dsgen_summary}
\end{table}

\begin{figure}[!htb] 
\centering
\includegraphics[width=0.5\textwidth, trim=0 0cm 0 0cm, clip]{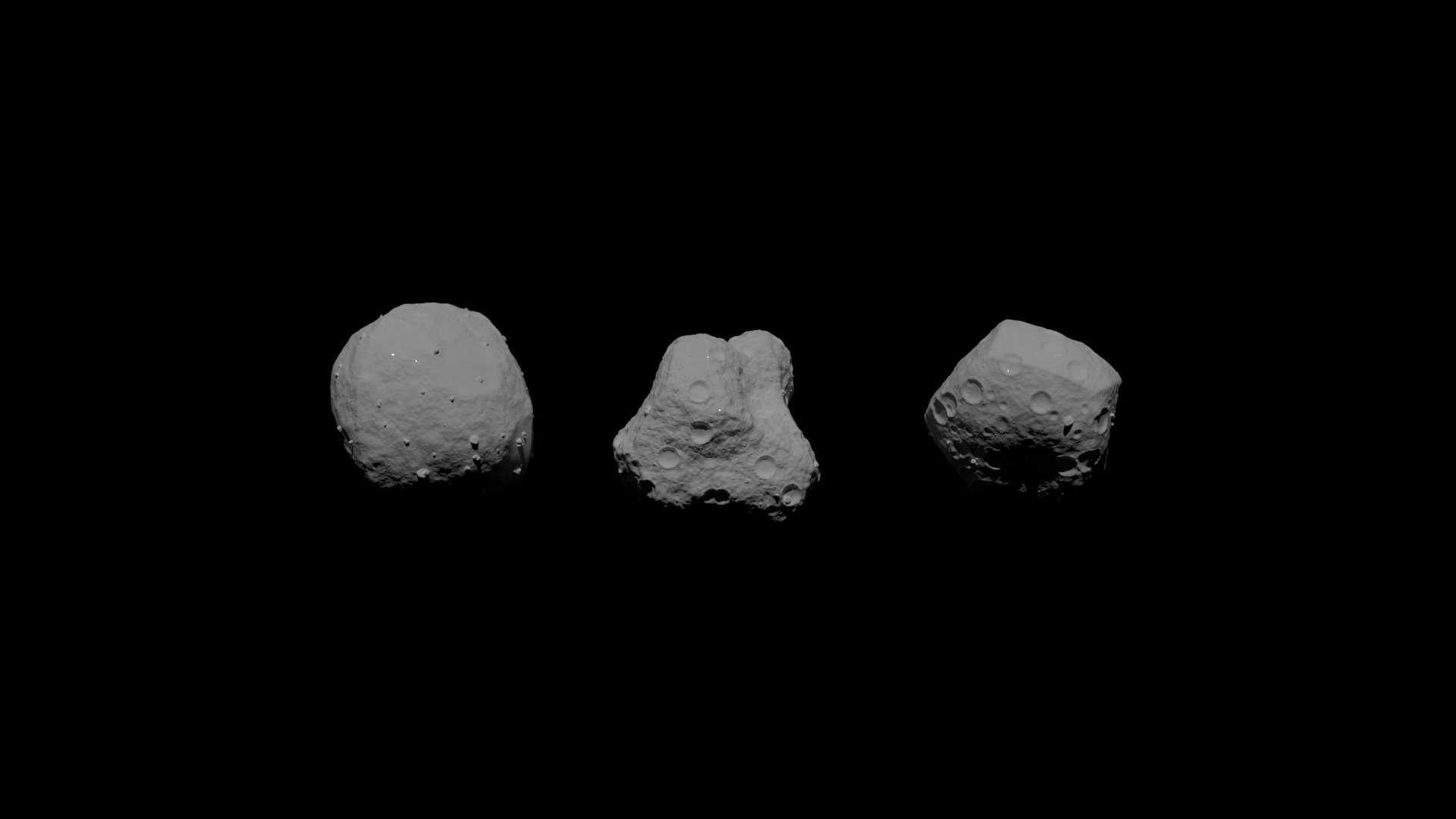} 
\caption{Renderings of three generated shape models (rendered with Blender)} 
\label{fig:generated_samples} 
\end{figure}

Figure \ref{fig:generated_samples} depicts renderings of three sample shape models drawn from the distribution in table \ref{table:dsgen_summary}.

For the benchmarking dataset, we used the shape model of the Apophis asteroid, reconstructed from radio observations during its 2014 fly-by of Earth \cite{brozowic2018}. Since the resolution of this shape model was limited, we manually added features to its surface and validated the resulting model against observations of other asteroids.
\begin{figure}[!htb] 
\centering
\includegraphics[width=0.5\textwidth, trim=0 0cm 0 0cm, clip]{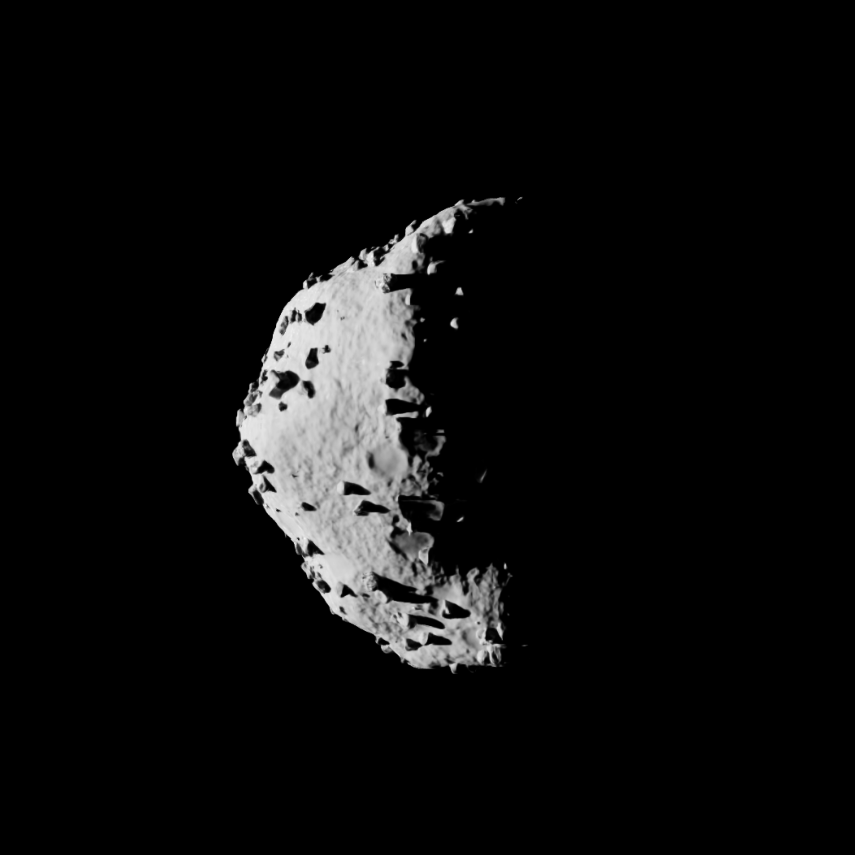} 
\caption{Renderings of the Apophis shape model (rendered with Nvidia Isaac Sim)} 
\label{fig:apophis} 
\end{figure}

Nvidia Isaac Sim and its ray-tracing renderer were finally used to generate frames of the shape models under different viewpoints. A random rotation was applied between subsequent frames for the train, validation and test datasets, whereas a tumbling motion was applied to the benchmarking shape model.

The dataset is finally composed of pairs of subsequent frames, along with their relative ground-truth pose and the depth map for each frame.

\section{Training}
\label{training}
Our feature description and feature matching models are aggregated into a common training pipeline.

An Adam optimizer with a $10^{-5}$ learning rate and a 0.95 exponential decay per epoch is used to update the weights of the model for 50 epochs. A single loss function is computed from the output of the feature matching model and gradients are propagated from the loss through the matching model to the description model.

The training data consists of pairs of images $I_A$ and $I_B$ linked by a random rotation $R_A^B$ of the asteroid shape model. Additionally, the ground-truth depth maps $D_A$ and $D_B$ are given.

To determine if the keypoints from $I_A$ and $I_B$ correspond to each other, we reproject the detected keypoints from $I_A$ to the reference frame of $I_B$ thanks to $R_A^B$ and $D_A$. If two keypoints are closer than one pixel in the same reference frame, they are said to correspond to one another. A ground-truth correspondence matrix $G$ is defined, such that each element $G_{ij}$ equals one if the feature $i$ in image A truly matches the feature $j$ in image B. Otherwise, it is zero.

The Lightglue feature matching model outputs a log-score match matrix $S$. Each element $S_{ij}$ of this matrix represents the log-probability that the feature $i$ of $I_A$ corresponds to the feature $j$ of $I_B$.

The loss function used to train our algorithm is the cross-entropy loss, which can be computed with the scores $S$ and the ground-truth correspondences $G$ such that:
$$
L = -\sum_i\sum _j S_{ij} G_{ij}
$$
Regularization of the loss function is achieved by introducing a uniform noise to the ground-truth correspondence matrix $G$. 

\section{Results}

COFFEE is compared against other state-of-the-art algorithms in terms of performance metrics, computational efficiency and accuracy of the full pose estimation framework. Each of the algorithms used to compare our algorithm against is tuned to perform optimally on our validation dataset, so that the bias in comparing the multiple algorithms is minimized. It is expected that the data-driven algorithms perform more accurately but less efficiently than their classical counterparts. With COFFEE, we aim to generally outperform the classical algorithms in terms of accuracy, while still having a higher computational efficiency than data-driven algorithms.

The algorithms against which COFFEE is benchmarked are the following:
\begin{enumerate}
\item SIFT (classical) \cite{sift}
\item ORB (classical) \cite{orb}
\item AKAZE (classical) \cite{akaze}
\item Superpoint (learned) \cite{superpoint}
\item ContextDesc (learned) \cite{contextdesc}
\item Disk (learned) \cite{disk}
\item LFNet (learned) \cite{lfnet}
\item R2D2 (learned) \cite{r2d2}
\end{enumerate}

After feature matching, a matrix corresponding to the match scores is established. A criterion, such as a threshold, a row-wise maximum or a K-best selection can be used to obtain the matrix of predicted matches $M$. Each element $M_{ij}$ equals one if the feature descriptor $i$ in image A is predicted to match the feature descriptor $j$ in image B. Otherwise, it is zero.

Having the ground-truth correspondence matrix $G$ and a matrix $M$ of predicted matches, we can compute the precision, recall and F1 score for the COFFEE description algorithm, as well as for the other baseline feature description algorithms. Tuning a feature description algorithm or the matching criterion typically increases the precision but decreases the recall or vice-versa. The F1-score overcomes this bias by introducing a metric that computes the harmonic mean between precision and recall. For completeness, the precision $P$, recall $R$ and F1-score $F_1$ are defined as follows:

\begin{equation} 
P = \frac{\sum_{i,j} G_{ij} M_{ij}}{\sum_{i,j} M_{ij}} 
\end{equation} 
\begin{equation} 
R = \frac{\sum_{i,j} G_{ij} M_{ij}}{\sum_{i,j} G_{ij}} 
\end{equation} 
\begin{equation} 
F_1 = 2 \cdot \frac{P \cdot \text{R}}{\text{P} + \text{R}} 
\end{equation}

For each algorithm, the K-best features are selected for multiple $K \in {100, 200, 500}$ and the measured precision is stated in table \ref{table:results_kbest_precision}. The precision after feature matching for a fixed number of features is a useful metric because only a constant number of matches are needed to estimate the pose of the tumbling asteroid. Note that the Non-Maximal Suppression algorithm \cite{sift} is applied, discarding matches that do not have a significantly higher score than the second-best matches. The ratio separating the best matches from the second-best matches is called the Lowe ratio and is tuned for each algorithm in our experiments. 

\begin{table}[!htb] 
\centering 
\begin{tabular}{|l|c|c|c|} 
\hline \textbf{Algorithm} & \textbf{100 features} & \textbf{200 features} & \textbf{500 features} \\ 
\hline COFFEE (ours) & \textbf{82.5\%} & \textbf{77.5\%} & \textbf{68.1\%} \\ 
\hline Superpoint & 69.4\% & 52.9\% & 30.4\%\\ 
\hline ContextDesc & 47.1\% & 45.1\% & 37.3\%\\ 
\hline Disk & 67.5\% & 57.4\% & 41.3\%\\ 
\hline LFNet & 16.2\% & 14.1\% & 10.7\%\\ 
\hline R2D2 & 16.9\% & 15.0\% & 10.9\%\\ 
\hline SIFT & 17.4\% & 14.8\% & 10.8\%\\ 
\hline ORB & 5.2\% & 4.3\% & 3.1\%\\ 
\hline AKAZE & 9.4\% & 8.0\% & 5.1\%\\ 
\hline \end{tabular} 
\caption{Precision for a given number of features} 
\label{table:results_kbest_precision}
\end{table}

\begin{figure}[!htb] 
\centering
\includegraphics[width=0.5\textwidth, trim=0 0cm 0 0cm, clip]{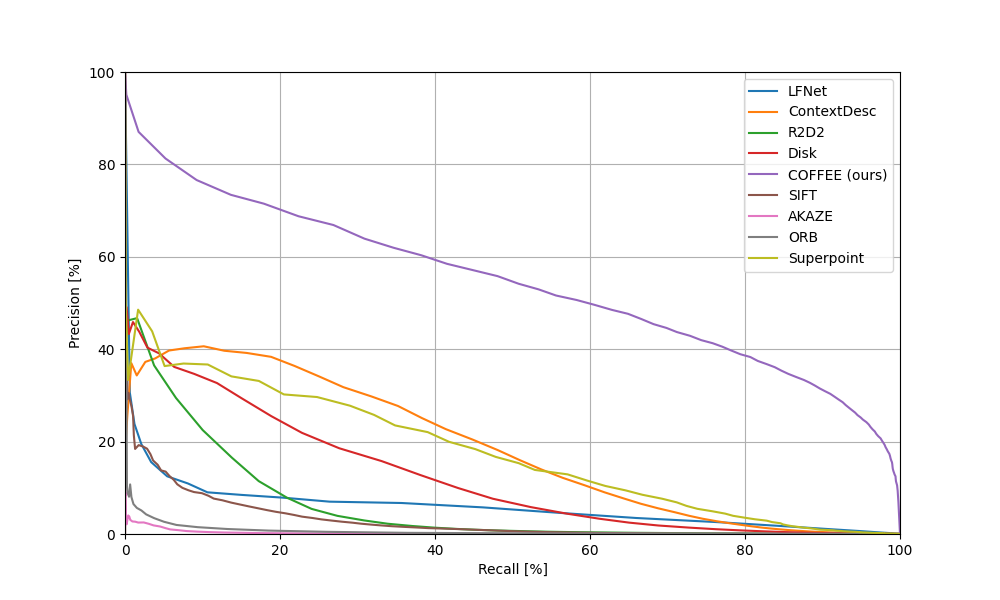} 
\caption{Precision-Recall curve for all benchmarked algorithms} 
\label{fig:pr_auc} 
\end{figure}

A threshold is applied to the feature matching algorithms, so that the trade-off between precision and recall can be analyzed. By sweeping the matching threshold, we plot the Precision-Recall curves for multiple algorithms in figure \ref{fig:pr_auc}. Additionally, the Area Under the Curve (AUC) produced by this sweep and the optimal F1-score are computed and reported in table \ref{table:auc}. \\
\begin{table}[!htb] 
\centering 
\begin{tabular}{|l|c|c|} 
\hline \textbf{Algorithm} & \textbf{Best F1-score} & \textbf{PR AUC} \\ 
\hline COFFEE (ours) & \textbf{54.7\%} & \textbf{54.2\%} \\ 
\hline Superpoint & 28.4\% & 17.6\% \\ 
\hline ContextDesc & 30.3\% & 18.4\% \\
\hline Disk & 20.8\% & 12.2\% \\ 
\hline LFNet & 9.6\% & 5.9\% \\ 
\hline R2D2 & 14.6\% & 5.9\% \\ 
\hline SIFT & 9.1\% & 2.8\% \\ 
\hline ORB & 3.5\% & 0.7\% \\ 
\hline AKAZE & 2.5\% & 0.2\% \\ 
\hline \end{tabular} 
\caption{ROC AUC, PR AUC and optimal F1-score for multiple algorithms} 
\label{table:auc}
\end{table}
We observe that COFFEE achieves a significantly higher precision, F1-score and PR AUC than any other algorithm, especially compared to the classical descriptors, such as SIFT, ORB and AKAZE. Nevertheless, the learned baseline descriptors faced a potentially significant domain-gap, as most of them have been trained on urban environments, whereas COFFEE was trained on our synthetic asteroid dataset. Eventually, we can only conclude that COFFEE performs well in absolute terms and relatively better than any classical descriptors, which are considered state-of-the-art in space missions.

\subsection{Pixel Error}
The quality of the pose estimation algorithm is directly related to the ability of the matching algorithm to accurately locate corresponding keypoints. The pixel error, defined as the median $L_2$ distance between matching keypoints, must be minimized.  \\

\begin{figure}[!htb] 
\centering
\includegraphics[width=0.5\textwidth, trim=0 0cm 0 0cm, clip]{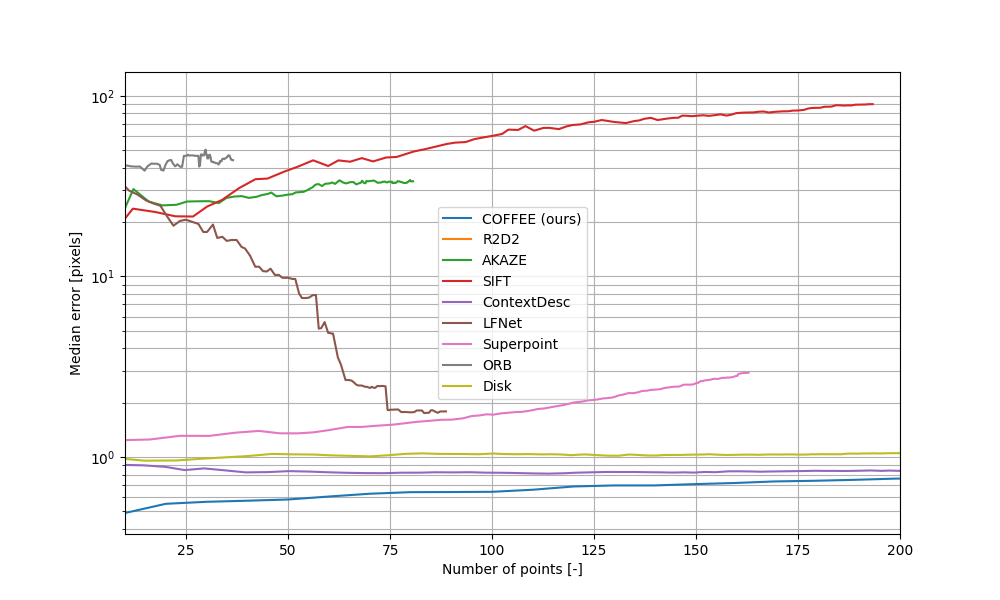} 
\caption{Median pixel error for multiple algorithms with a Lowe ratio of 0.9} 
\label{fig:pixel_error_lowe}
\end{figure}

We compare the baseline algorithms with our algorithm by plotting the pixel error against the number of matched keypoints in figure \ref{fig:pixel_error_lowe}. Matches are ranked and selected based on their matching score.
The median pixel error of COFFEE then becomes comparable to the Disk \cite{disk} and ContextDesc \cite{contextdesc} deep learning feature descriptors at a subpixel accuracy. At a low number of points, the Superpoint \cite{superpoint} also achieves subpixel accuracy. We conclude that these four algorithms will likely be the most suitable for the pose estimation algorithm and that COFFEE yields a significantly lower error than all classical algorithms for any number of correspondences, thanks to its robustness to the motion of shadows.

\subsection{Pose Estimation Error}
We estimate the pose of the asteroid by applying a RANSAC scheme to the 5-point algorithm, recovering the essential matrix and inferring the relative rotation matrix and pose, as explained in section \ref{poseest}. 
The resulting estimation errors are stated in tables \ref{table:error100} and \ref{table:error2000} for the best 100 and 2000 features respectively.
\begin{table}[!htb] 
\centering 
\begin{tabular}{|l|c|c|} 
\hline \textbf{Algorithm} & \textbf{Error [rad]}  & \textbf{Standard deviation [rad]}\\ 
\hline COFFEE (ours) & \textbf{0.029} & \textbf{0.018} \\ 
\hline Superpoint & 0.041 & 0.020\\ 
\hline ContextDesc & 0.13 & 0.12\\
\hline Disk & 0.20 & 0.29 \\ 
\hline LFNet & 0.62 & 0.85 \\ 
\hline R2D2 & 0.12 & 0.089 \\ 
\hline SIFT & 0.29 & 0.37 \\ 
\hline ORB & 0.81 & 0.96 \\ 
\hline AKAZE & 0.41 & 0.48 \\ 
\hline \end{tabular} 
\caption{Pose estimation bias and std dev. for the best K=100 features} 
\label{table:error100}
\end{table}
\begin{table}[!htb] 
\centering 
\begin{tabular}{|l|c|c|} 
\hline \textbf{Algorithm} & \textbf{Error [rad]}  & \textbf{Standard deviation [rad]}\\ 
\hline COFFEE (ours) & \textbf{0.034} &0.027 \\ 
\hline Superpoint & 0.038 & \textbf{0.026}\\ 
\hline ContextDesc & 0.050 & 0.037\\
\hline Disk & 0.088 & 0.080 \\ 
\hline LFNet & 0.068 & 0.066 \\ 
\hline R2D2 & 0.061 & 0.046 \\ 
\hline SIFT & 0.12 & 0.099 \\ 
\hline ORB & 0.41 & 0.54 \\ 
\hline AKAZE & 0.20 & 0.12 \\ 
\hline \end{tabular} 
\caption{Pose estimation bias and std dev. for the best K=2000 features} 
\label{table:error2000}
\end{table}

With a low number of features ($K=100$), COFFEE  achieves a bias error less than $0.03$ [rad] and a standard deviation less than $0.02$ [rad], a comparable result to some deep learning methods such as Superpoint. When increasing $K$ to 2000, the performance gap with the baseline algorithm is reduced, even though COFFEE retains a low bias error and standard deviation. Note that having such a high number of features is limited by the surface roughness of the target asteroid and the resolution of the onboard camera.

In both cases, the low bias error exhibited by COFFEE demonstrates that the latter is robust to the effect of moving shadows, as opposed to the other benchmarked algorithms. To support this claim, we created renderings of a fictitious asteroid with little surface roughness and large features (craters and boulders). 

\begin{figure}[!htb] 
\centering
\includegraphics[width=0.5\textwidth, trim=0 0cm 0 0cm, clip]{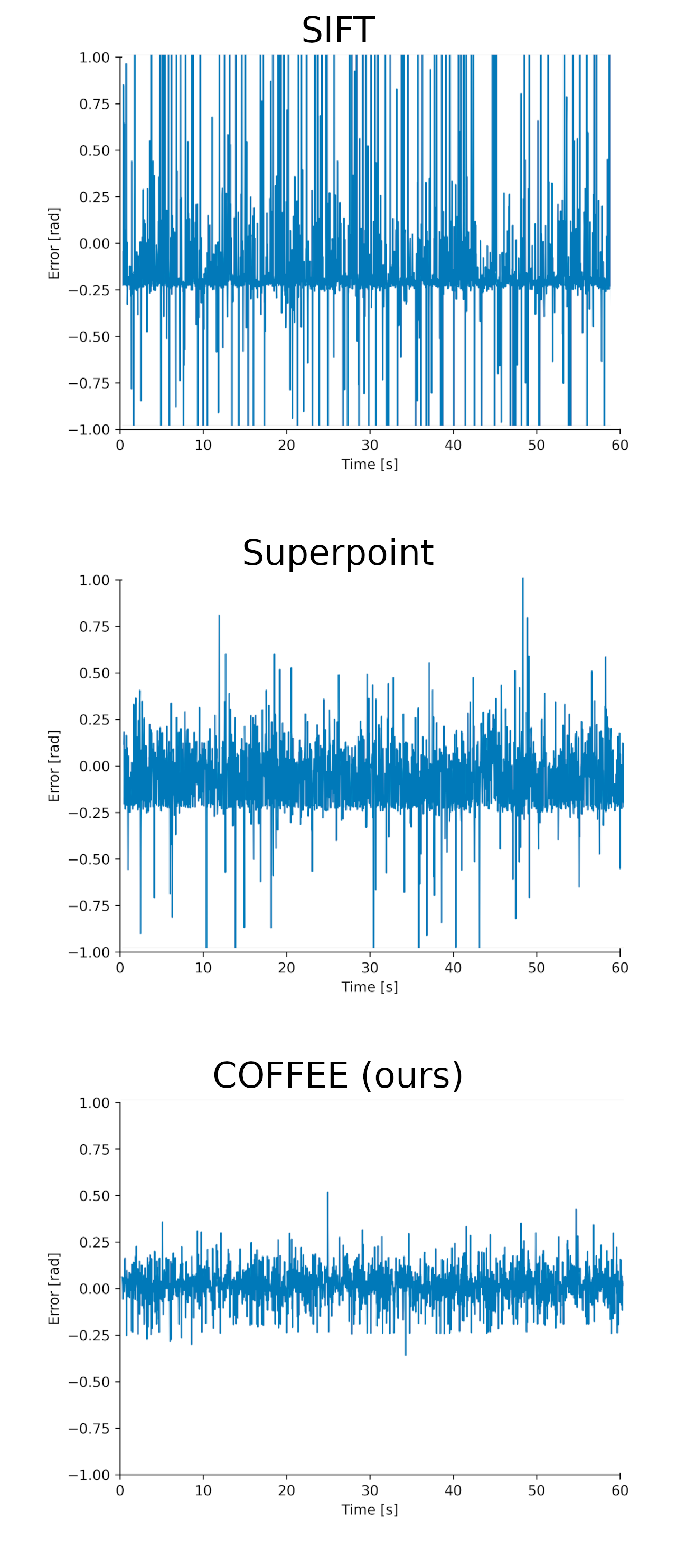} 
\caption{Pose estimation bias for multiple algorithms} 
\label{fig:comparison_bias}
\end{figure}

We apply SIFT, Superpoint and COFFEE on this dataset to estimate the pose of the rotating asteroid and plot the resulting errors in figure \ref{fig:comparison_bias}. 

As expected, our method yields a bias-free pose estimation, as opposed to classical feature descriptors, like SIFT, or data-driven descriptors, like Superpoint. In addition to this, the variance of the error is much lower than with SIFT and Superpoint, which supports the evidence stated in table \ref{table:error100} and \ref{table:error2000}. The low number of visible features on this fictitious asteroid increases the error gap between COFFEE and the benchmarked algorithms.

\subsection{Runtime Metrics}
Although parallel computing hardware, in particular GPUs, have seen a swift increase in performance and memory capacity in recent years, space hardware remains years behind the current state-of-the-art. This is mainly due to the conservative requirements for space hardware, the slow process of verifying and validating these requirements, as well as the harsh conditions of the space environment. Candidate pose estimation algorithms must consequently evaluate at a rate high enough to track the pose of the target asteroid, especially if the latter is tumbling. As this estimation rate is determined by the mission requirements, we show here that COFFEE comparatively performs better than the classical pipelines and faster than data-driven pipelines.

COFFEE is compared against the baseline descriptors by running the algorithms on the same desktop computer with a Nvidia Titan RTX GPU and a 20-cores Intel Core i9-7900X.

\begin{figure}[!htb] 
\centering
\begin{tikzpicture}
    \begin{axis}[
        xmode=log,
        width=8cm,
        height=8cm,
        xlabel={Framerate (fps)},
        ylabel={Error (rad)},
        grid=major,
        scatter/classes={
            a={mark=*,draw=black,fill=orange},
            b={mark=triangle*,draw=black,fill=red},
            c={mark=square*,draw=black,fill=blue},
            d={mark=diamond*,draw=black,fill=green},
            e={mark=pentagon*,draw=black,fill=yellow},
            f={mark=*,draw=black,fill=olive},
            g={mark=triangle*,draw=black,fill=purple},
            h={mark=square*,draw=black,fill=brown},
            i={mark=diamond*,draw=black,fill=blue}},
        legend pos=north west
    ]

    \addplot[scatter,only marks,
        scatter src=explicit symbolic] 
        table[meta=class] {
        x y class label
        70.4 0.029 a
        19.1 0.038 b
        2.5 0.050 c
        10.3 0.088 d
        22.7 0.068 e
        2.0 0.061 f
        82 0.12 g
        227 0.41 h
        179 0.20 i
    };
    \legend{COFFEE (ours), Superpoint, ContextDesc, Disk, LFNet, R2D2, SIFT, ORB, AKAZE}

    \end{axis}
\end{tikzpicture}
\caption{Trade-off between runtime and accuracy}
\label{fig:tradeoff}
\end{figure}
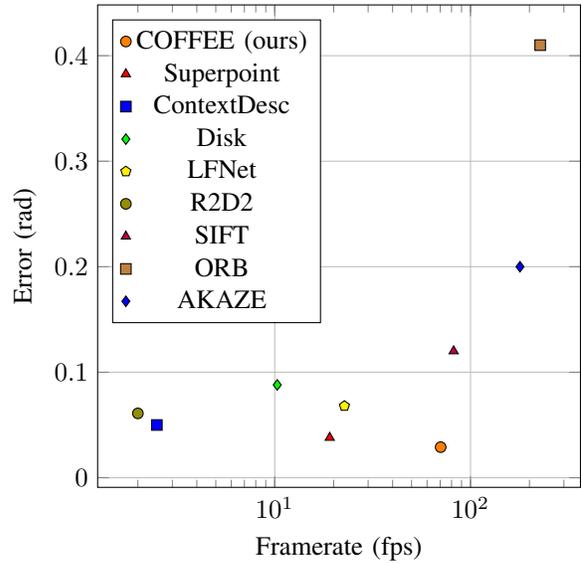

The trade-off between overall pose estimation accuracy and runtime in figure \ref{fig:tradeoff} shows that the COFFEE algorithm is much faster than the other deep learning algorithms but much more accurate than the other classical algorithms. We have effectively pushed the pareto optimum of the runtime-accuracy trade-off and designed an algorithm that leverages the advantages of both classical and learned algorithms to obtain a lightweight small body pose estimation pipeline for space applications.

\subsection{Qualitative Comparisons}
\begin{figure}[!htb]
\centering
\includegraphics[width=0.5\textwidth, trim=0 0cm 0 0cm, clip]{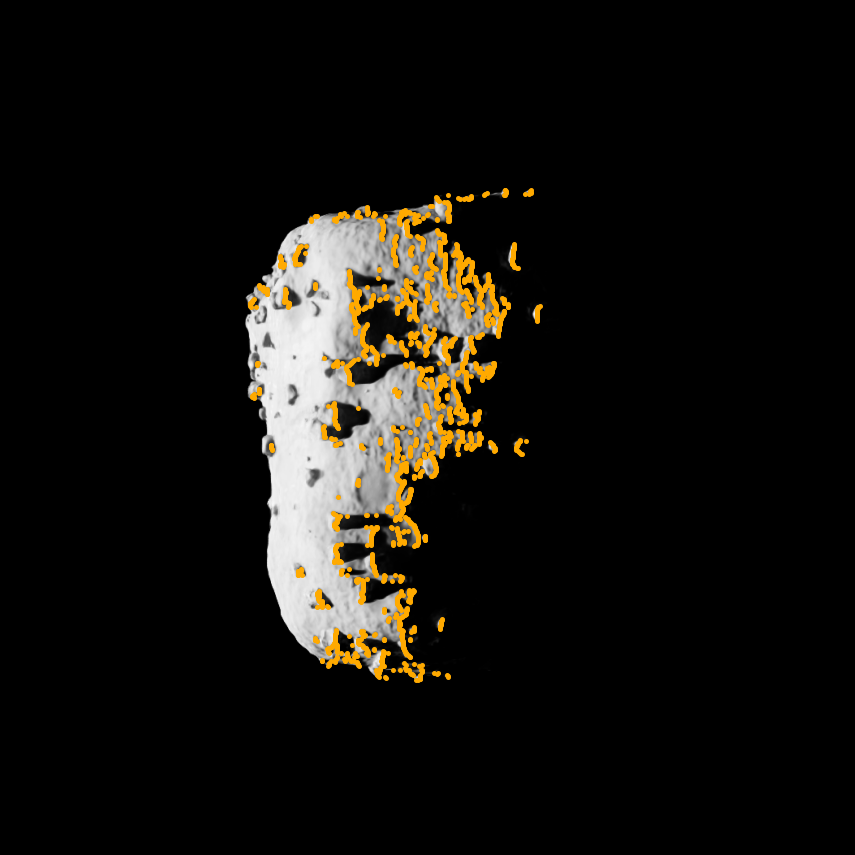} 
\caption{Keypoints detected by the COFFEE algorithm} (orange keypoints) 
\label{fig:detection}	
\end{figure}

Figure \ref{fig:detection} shows renderings of the Apophis asteroids, under a pure rotation around the vertical axis (with respect to the camera frame). Additionally, the sun phase angle is 90°, implying that the vanishing point is infinitely to the east of the image, and that all shadows are cast from the left to the right in the camera’s reference frame. This worst-case is carefully analyzed, as it is key to the validation of the robustness of the proposed COFFEE algorithm.

The keypoints detected by COFFEE are highlighted in orange. Its detector extracts information about the shadow by analyzing the projection of the rays emanating from the sun. 

\begin{figure}[!htb] 
\centering
\includegraphics[width=0.5\textwidth, trim=0 0cm 0 0cm, clip]{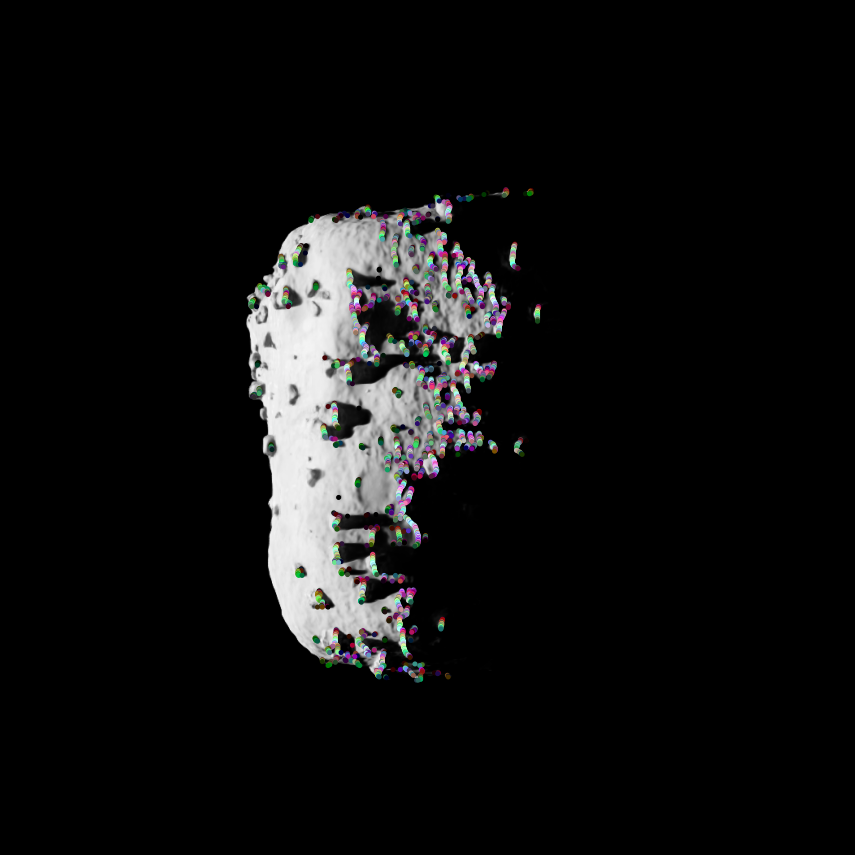} 
\caption{Features described by the COFFEE algorithm (colored keypoints)}
\label{fig:description}
\end{figure}

Figure \ref{fig:description} shows the assignment of a feature descriptor to each of the detected keypoints, thanks to the Sparse Submanifold ResNet model. To ease visualization, we projected the resulting 256-dimensional vector to a 3-dimensional color space through a random projection matrix.

\begin{figure}[!htb]
\centering
\includegraphics[width=0.5\textwidth, trim=0 0cm 0 0cm, clip]{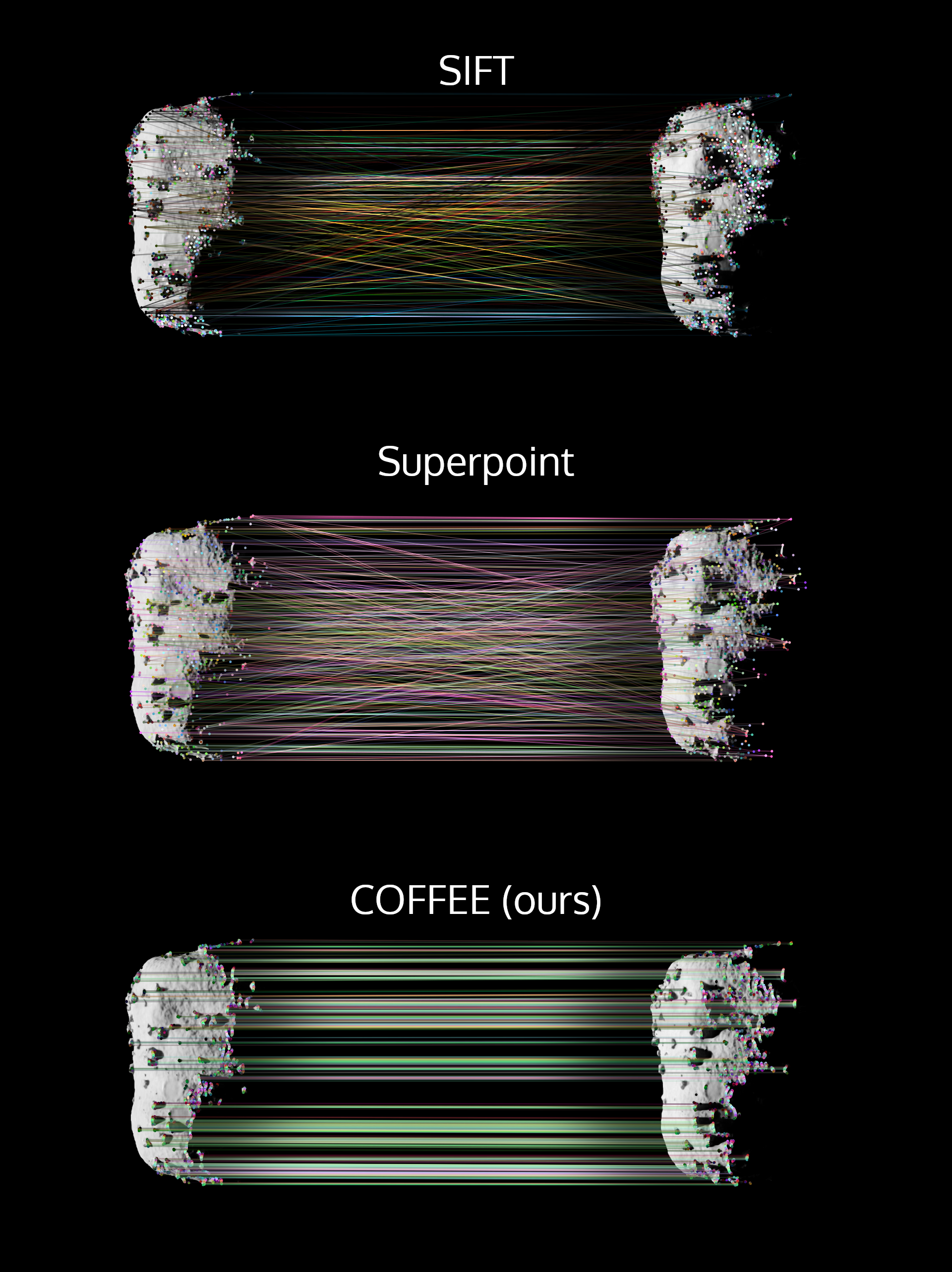} 
\caption{Qualitative comparison of matched features by the SIFT, Superpoint and COFFEE algorithms}
\label{fig:matched}
\end{figure}

Figure \ref{fig:matched} depicts the keypoint correspondences between subsequent frames that SIFT, Superpoint and COFFEE found for a fixed number of matches. Since the body is rotating along the vertical axis, we should qualitatively see all matches lying horizontally in the image.

We observe that the quality of correspondences found by COFFEE is superior to the quality of those found by Superpoint and SIFT.

\section{Conclusion}
By carefully analyzing the effect of self-cast shadows on asteroids, we found an efficient algorithm to extract shadow-resilient keypoints from images taken onboard a spacecraft. The 3D geometry of the asteroid, embedded in a 2D shadow and extracted as a 1D boundary proved to be sufficiently informative to recover the pose of the asteroid. To this end, we designed a deep learning model to efficiently capture all information about this boundary. Using a Residual Network architecture, implemented as a Submanifold Sparse CNN, we could describe each extracted keypoint by a unique 256-dimensional vector. An attention-based GNN was then used to find feature correspondences between the two images. A robust RANSAC scheme was finally applied with the 5-points algorithm to infer the relative pose of the asteroid between subsequent frames. 

In addition to qualitative results, we provided quantitative metrics showing that our algorithm performed better than classical algorithms in terms of precision, recall, $F_1$ score, Precision-Recall Area-Under-Curve, and especially pose estimation bias. Moreover, we showed that COFFEE runs at least 3x faster than the benchmarked deep learning algorithms. 

We nevertheless acknowledge that the large domain gap faced by the baseline data-driven algorithms could have slightly skewed the results in our favor, yet without affecting any conclusion regarding the higher computational efficiency of COFFEE. We believe that the absolute accuracy in pose estimation and the very fast runtime of our method could make it a good candidate for future asteroid missions during their orbit synchronization and final approach phase. 

This research could be extended by exploring the benefits of online domain adaptation for the pose estimation pipeline and integrating the latter within a complete Simultaneous Localization and Mapping (SLAM) system to increase the robustness and accuracy of our method. Moreover, COFFEE could be integrated in a complete vision-in-the-loop control system, so that the robustness and efficiency of our method would benefit the whole navigation stack of the spacecraft.

\pagebreak

\bibliographystyle{ieeetr}
\bibliography{paper.bib}

\clearpage
\onecolumn
\appendix
\begin{figure}[!htb]
\centering
\includegraphics[width=\textwidth, trim=0 0cm 0 0cm, clip]{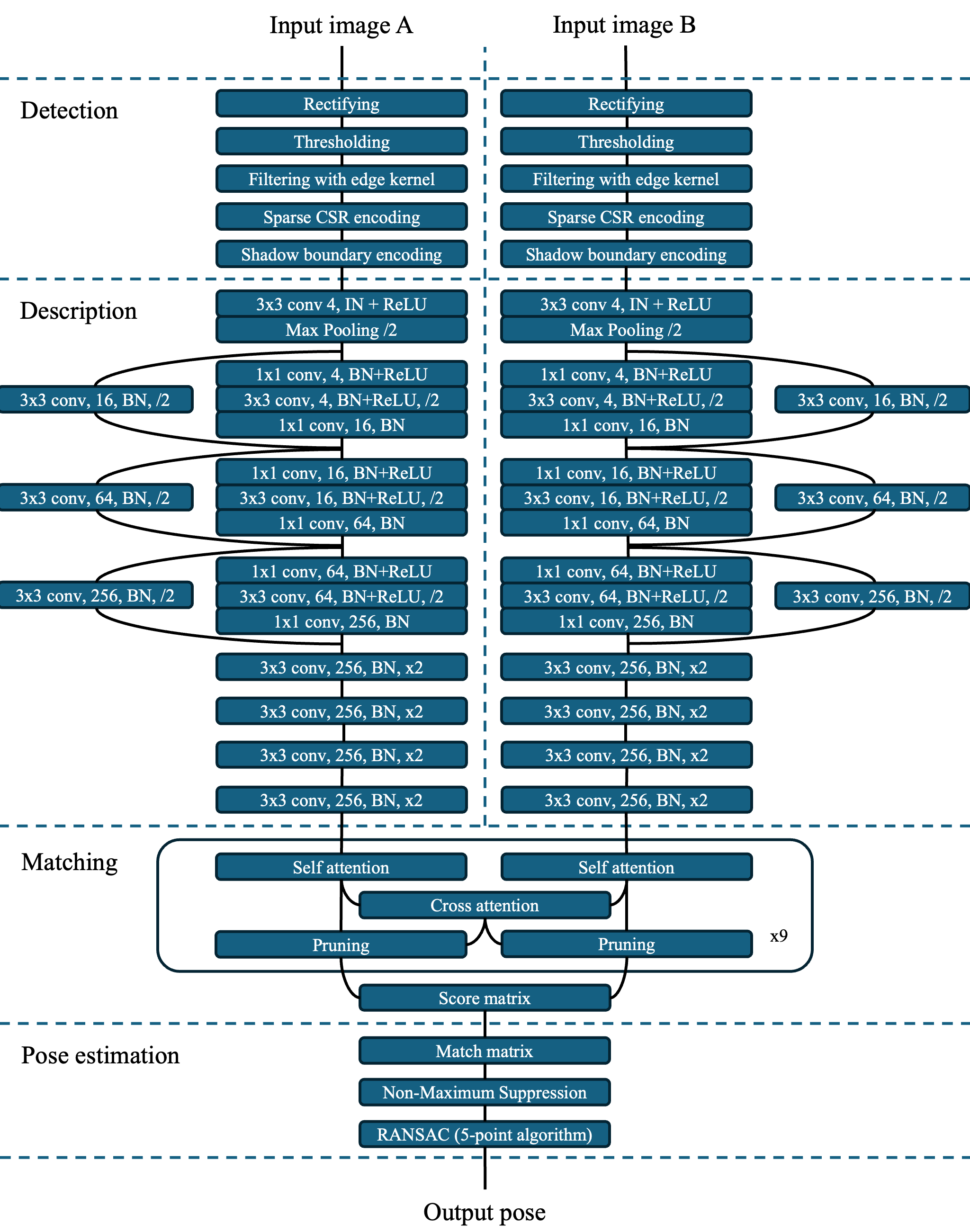} 
\caption{Summarized architecture of the COFFEE pipeline}
\label{fig:pipeline}
\end{figure}

\end{document}